\documentclass[pdflatex,sn-mathphys-num]{sn-jnl}
\usepackage{amsmath,amssymb,amsfonts}%
\usepackage{algorithm}
\usepackage{algpseudocode}
\usepackage{array}
\usepackage{textcomp}
\usepackage{stfloats}
\usepackage{url}
\usepackage{verbatim}
\usepackage{graphicx}

\usepackage{hyperref}

\usepackage{subcaption}
\usepackage{multirow}
\usepackage{balance}
\usepackage{xspace}
\usepackage{xcolor}
\usepackage{booktabs}
\usepackage [english]{babel}
\usepackage [autostyle, english=american]{csquotes}
\MakeOuterQuote{"}
\usepackage{tabularx}
\usepackage{array}
\newcolumntype{C}{>{\centering\arraybackslash}X}

\newcolumntype{H}{>{\setbox0=\hbox\bgroup}c<{\egroup}@{}}

\usepackage{xstring}
\usepackage{graphicx}
\usepackage{etoolbox}
\usepackage{colortbl}
\usepackage{fontawesome5} 

\def\tick{\faCheckCircle}

\def\myalgoname{\texttt{pi-CE-VAE}\@\xspace}
\def\myGAN{\texttt{GAN}\@\xspace}
\def\myRbA{\texttt{RbA}\@\xspace}

\def\conv{\texttt{Conv}}
\def\convtwod{\texttt{Conv2D}}
\def\convT{\texttt{TConv2D}}

\def\concat{\texttt{Concat}}

\def\upsample{\texttt{UpSample}}
\def\resblock{\texttt{ResnetBlock}\xspace}
\def\upsampleblock{\texttt{UpSampleBlock}\@\xspace}
\def\avgpool{\texttt{AvgPool}}

\def\encoder{E}
\def\decoder{D}
\def\capsuleclustering{Q}
\def\physicsestimator{\Phi}
\def\physicsenhancer{\Psi}

\newcommand{\tblfirst}[1]{\textbf{\textcolor{red}{#1}}}
\newcommand{\tblsecond}[1]{\textbf{\textcolor{blue}{#1}}}

\def\image{\mathbf{I}}
\def\inputimg{\image_{\text{deg}}}
\def\reconstructedinputimg{\widehat{\image}_{\text{deg}}}

\def\realcleanoutputimg{\image_{\text{clear}}}
\def\reconstructedimg{\widehat{\image}_\text{clear}}
\def\reconstructedintermediateimg{\Tilde{\image}_{\text{clear}}}

\def\transmissionmap{\mathbf{T}}
\def\transmissionmapestimate{\widehat{\transmissionmap}}
\def\backgroundlight{\alpha}
\def\backgroundlightestimate{\widehat{\backgroundlight}}

\def\loss{\mathcal{L}}

\def\ie{\textit{i.e.}\@\xspace}
\def\eg{\textit{e.g.}\@\xspace}

\usepackage{graphicx}
\usepackage{etoolbox}

\newcommand{\insertfig}{\setcounter{figure}{0}\captionsetup{type=figure}\includegraphics[width=\linewidth]{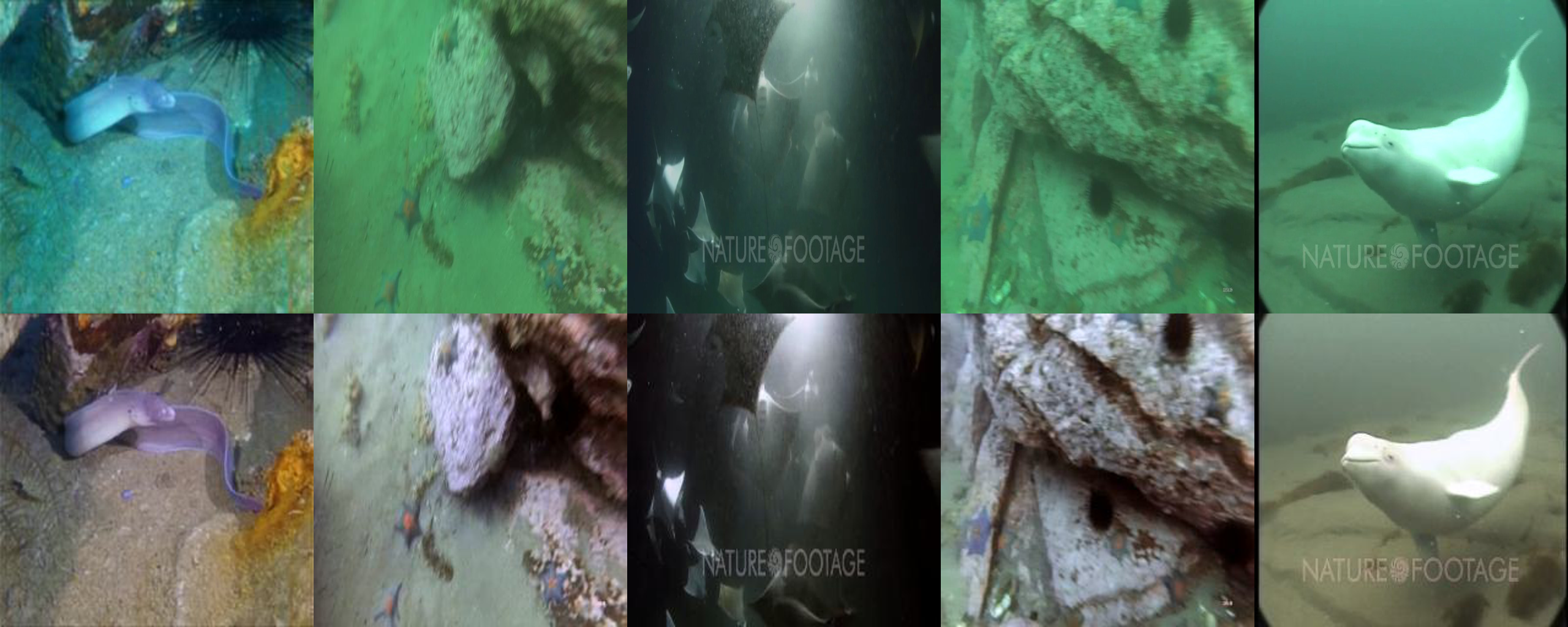}\captionof{figure}{Results of our approach for underwater compressed image reconstruction and enhancement. The first row shows the effects of water light refraction in underwater images introducing blurriness and cold greenish or bluish tone (among others) issues. The second row shows the enhanced results obtained by the proposed approach.}\label{fig:teaser}
}

\makeatletter
\apptocmd{\@maketitle}{\centering\insertfig}{}{}

\def\subsubsection{\@startsection{subsubsection}{3}{\parindent}{1ex plus 0.1ex minus 0.1ex} %
{0ex plus .0ex minus 0ex}{\normalfont\normalsize\itshape\underline}}%

\makeatother

\begin{document}
\title{Physics Informed Capsule Enhanced Variational AutoEncoder for Underwater Image Enhancement}

\author*[1]{\fnm{Niki} \sur{Martinel}}\email{niki.martinel@uniud.it}
\author[2]{\fnm{Rita} \sur{Pucci}}\email{puccirp@vuw.leidenuniv.nl}

\affil*[1]{\orgdiv{Department of Mathematics, Computer Science and Physics}, \orgname{University of Udine}, \orgaddress{\street{Via Delle Scienze 206}, \city{Udine}, \postcode{33100}, \country{Italy}}}

\affil[2]{
\orgdiv{Leiden Institute of Advanced Computer Science}, 
\orgname{University of Leiden}, 
\orgaddress{\street{Einsteinweg 55},\city{Leiden}, 
\postcode{2333},\country{The Netherlands}}}



\abstract{
We present a novel dual-stream architecture that achieves state-of-the-art underwater image enhancement by explicitly integrating the Jaffe-McGlamery physical model with capsule clustering-based feature representation learning.
Our method simultaneously estimates transmission maps and spatially-varying background light through a dedicated physics estimator while extracting entity-level features via capsule clustering in a parallel stream. This physics-guided approach enables parameter-free enhancement that respects underwater formation constraints while preserving semantic structures and fine-grained details.
Our approach also features a novel optimization objective ensuring both physical adherence and perceptual quality across multiple spatial frequencies.
To validate our approach, we conducted extensive experiments
across six challenging benchmarks.
Results demonstrate consistent improvements of $+0.5$dB PSNR over the best existing methods while requiring only one-third of their computational complexity (FLOPs), or alternatively, more than $+1$dB PSNR improvement when compared to methods with similar computational budgets.
Code and data \textit{will} be available at \url{https://github.com/iN1k1/}.
}

\maketitle


\section{Introduction}
Underwater imaging presents unique challenges that distinguish it from traditional computer vision applications. The aquatic medium introduces complex degradation phenomena, including wavelength-dependent absorption, forward and backward scattering, and spatially varying illumination conditions resulting in images characterized by color distortion, reduced contrast, and attenuated visibility (\eg,~\figurename~\ref{fig:teaser}, first row).
These degradations harm human visual perception and compromise the performance of downstream computer vision algorithms, making underwater image enhancement a fundamental preprocessing step for marine robotics~\cite{bingham2010robotic}, underwater surveillance~\cite{shkurti2012multi}, and oceanic exploration applications~\cite{whitcomb2000advances}.

The fundamental challenge in underwater image enhancement (UIE) lies in the inherent complexity of the image formation process. Unlike atmospheric imaging --where degradation models are relatively well-established-- underwater environments show highly dynamic and spatially-varying degradation patterns.
The Jaffe-McGlamery model~\cite{jaffe1990computer,mcglamery1975computer} provides a principled physical framework for understanding underwater image formation, describing how clear images are corrupted through direct transmission and backscattering components. However, translating this theoretical understanding into computational solutions remains non-trivial, as traditional approaches often struggle to accurately estimate the physical parameters while also preserving semantic content and fine-grained details.

Recent advances in UIE exploited (i) traditional image processing techniques and (ii) machine learning-based methods.
The former category includes non-physics-based~\cite{li2016underwater,ghani2015underwater} and physics-based~\cite{han2017active, neumann2018fast,hu2021underwater} approaches.
Due to their high parameterization characteristics --often requiring a detailed knowledge of the environment-- these methods lack generalization across diverse underwater settings.
Methods in the latter category mostly rely on deep learning architectures that lack explicit physical grounding.
While these methods can achieve visually appealing results~\cite{islam2020fast,park2019adaptive,hu2021underwater,fabbri2018enhancing,zhang2021dugan,zhu2017unpaired,islam2020simultaneous,guo2019underwater}, they may introduce artifacts that violate fundamental physical principles. 

\emph{To address these limitations, we propose a novel physics-informed UIE framework that integrates explicit image formation modeling with advanced deep learning architectures.}
Our key insight is that effective UIE requires both adherence to physical constraints and sophisticated feature representation learning.
We achieve this through a dual-stream architecture where one pathway focuses on estimating physical parameters (transmission maps and background light) while a parallel stream performs hierarchical feature extraction augmented by capsule clustering for entity-level representation learning.

Our physics estimator predicts spatially-varying transmission maps and background light distributions, enabling explicit modeling of the underwater degradation process.
These estimates are then leveraged by a parameter-free physics-informed enhancer that applies the inverse transformation according to the Jaffe-McGlamery model, ensuring that the enhancement process respects fundamental image formation principles.
The feature extraction stream captures semantic and structural information essential for perceptually compelling results, with capsule clustering providing entity-level representations that preserve spatial hierarchies and part-whole relationships.

To optimize our novel model, we introduce multiple complementary loss terms designed to enforce physical consistency and perceptual quality.
The former is achieved by cycle consistency and transmission supervision losses that guarantee adherence to the underlying physical model. 
The latter leverages a multi-scale pyramid loss ensure spatial coherence and multi-frequency detail preservation.

The primary contributions of this work are threefold:
\begin{itemize}
    \item We introduce a novel dual-stream architecture that explicitly integrates underwater physics modeling with advanced feature representation learning through capsule clustering;
    \item We propose an optimization objective that optimizes physical parameter estimation and perceptual enhancement quality across multiple spatial scales; 
    \item Through a compelling set of experiments on 6 benchmark datasets, we demonstrate state-of-the-art performance at a lower computational cost than current best performing solutions.
\end{itemize}

Our approach represents a significant step toward physics-aware underwater image enhancement that combines theoretical rigor with practical effectiveness.

\section{Related Works}
\label{sec:rw}
A recent survey~\cite{ma2025comparative} of underwater image enhancement methods classifies exiting solutions distinguishing between traditional and machine learning-based approaches. 
We follow the same principle to analyze the literature.

\emph{Traditional methods} focus on the estimation of global background and water light transmission to perform image enhancement.
In~\cite{bazeille2006automatic,8058463}, independent image processing steps have been proposed to correct non-uniform illumination, suppress noise, enhance contrast, and adjust colors.
Other methods introduced edge detection operations to implement object-edge preservation during filtering operations for color enhancement~\cite{lu2013underwater}.
In~~\cite{li2016single}, it has been observed that the image channels are affected differently by the disruption of light: red colors are lost after a few meters from the surface while green and blue are more persistent.
These differences introduced enhancement methods that act differently on each color channel and sacrifice generalization in favor of ad-hoc filters based on environmental parameters~\cite{park2017enhancing, 9744022}.
Other approaches estimated the global background light parameters~\cite{park2017enhancing, peng2017underwater} to apply specific color corrections (\ie, to reduce the blueish and greenish effects).
More recently, there has been a surge of interest in exploring the physics behind the Jaffe-McGlamery formation model for image restoration.
Approaches in this direction worked on contrast optimization~\cite{lin2025modified}, focused on diverse underwater environments by proposing context-aware solutions~\cite{wang2024lightweight}, or disentangle~\cite{yan2025underwater} the scattering components from the transmission component, also through depth map estimation and backscatter elimination~\cite{zhou2023}.
These models use the principles of light and color physics to account for various underwater conditions. Despite being more accurate, their application is limited due to the challenges of obtaining all the necessary variables that impact underwater footage.
Efforts have been made to improve the estimation of the global background light~\cite{akkaynak2019sea} at the cost of increasing algorithm complexity and overfitting experimental data with poor generalization on new test data.

\emph{Machine learning-based methods} for underwater image enhancement made extensive use of a U-Net-like structure~\cite{ronneberger2015u} to enhance the input image while preserving the spatial information and relationship between objects.
Skip connections are often used to propagate the raw inputs to the final layers to preserve spatial relationships~\cite{8917818,9986534} also with special attention and pooling layers~\cite{qiao2022adaptive}.
Other methods explored the emerging application of Transformers via channel-wise and spatial-wise attention layers~\cite{peng2023ushape-lsuidataset} or through customized transformer blocks leveraging both the frequency and the spatial-domains as self-attention inputs~\cite{khan2024spectroformer}.
Generative Adversarial Networks (\myGAN{}s) training schemes have also been explored for the task~\cite{guo2019underwater} along with 
approaches improving the information transfer between the encoder and decoder via multiscale dense blocks~\cite{islam2020fast} or hierarchical attentions modules~\cite{han2023fe}.
More recently, frequency- and diffusion-based strategies have emerged.
Diffusion-based enhancement using non-uniform skip strategy was introduced in~\cite{tang2023dmwater}, later extended by combining wavelet and Fourier transforms with a residual diffusion adjustment mechanism~\cite{zhao2024wfdiff} or by incorporating underwater physical priors to better guide image restoration~\cite{zhao2024padiff}.

We extend our preliminary results~\cite{pucci2025cevae} by a method that falls in the latter category while introducing novel model components that leverage the physics of the Jaffe-McGlamery image formation model.
While these machine-learning based methods achieve compelling results, they typically involve high computational overhead due to iterative sampling or heavy global attention mechanisms.
In contrast, our method introduces a novel dual-stream architecture that integrates the Jaffe-McGlamery physical image formation model with a capsule clustering-based feature representation.
This design removes the need for global attention mechanisms to properly model entity presences and location, enabling both enhancement and reconstruction from a lightweight, compressed representation.
Our dedicated physics estimator learns to predict transmission maps and spatially-varying background light, ensuring adherence to underwater imaging constraints via dedicated losses.
Capsule clustering extracts entity-level semantics, preserving fine-grained detail without relying on full-resolution context.
We demonstrated the benefits over such methods (\eg, more than $+1$ dB PSNR under similar computational constraints, or $+0.5$ db PSNR with one-third the FLOPs) through a throughout comparison on six existing benchmarks.

\begin{figure}[!t]
  \centering
  \includegraphics[width=\linewidth]{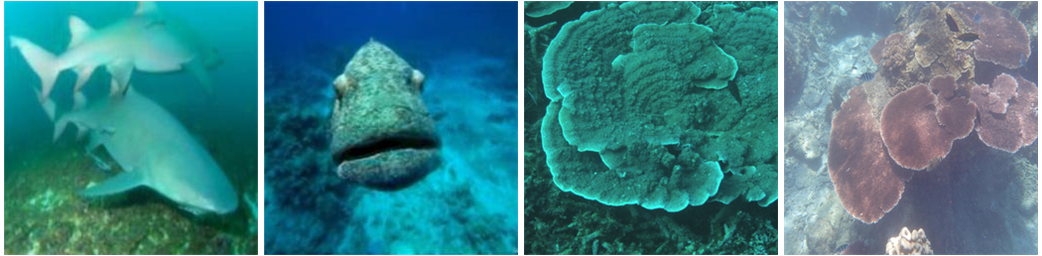}
  \caption{Samples shot in an underwater environment showing some typical underwater imaging issues  (left to right): low edge definition, blueish colors, greenish colors, and blurriness.
  Images are from public benchmark datasets.
  }
  \label{fig:UWIN}
\end{figure}

\section{Background} 
\label{sec:bg}
\begin{figure}[!t]
  \centering
  \includegraphics[width=0.6\linewidth]{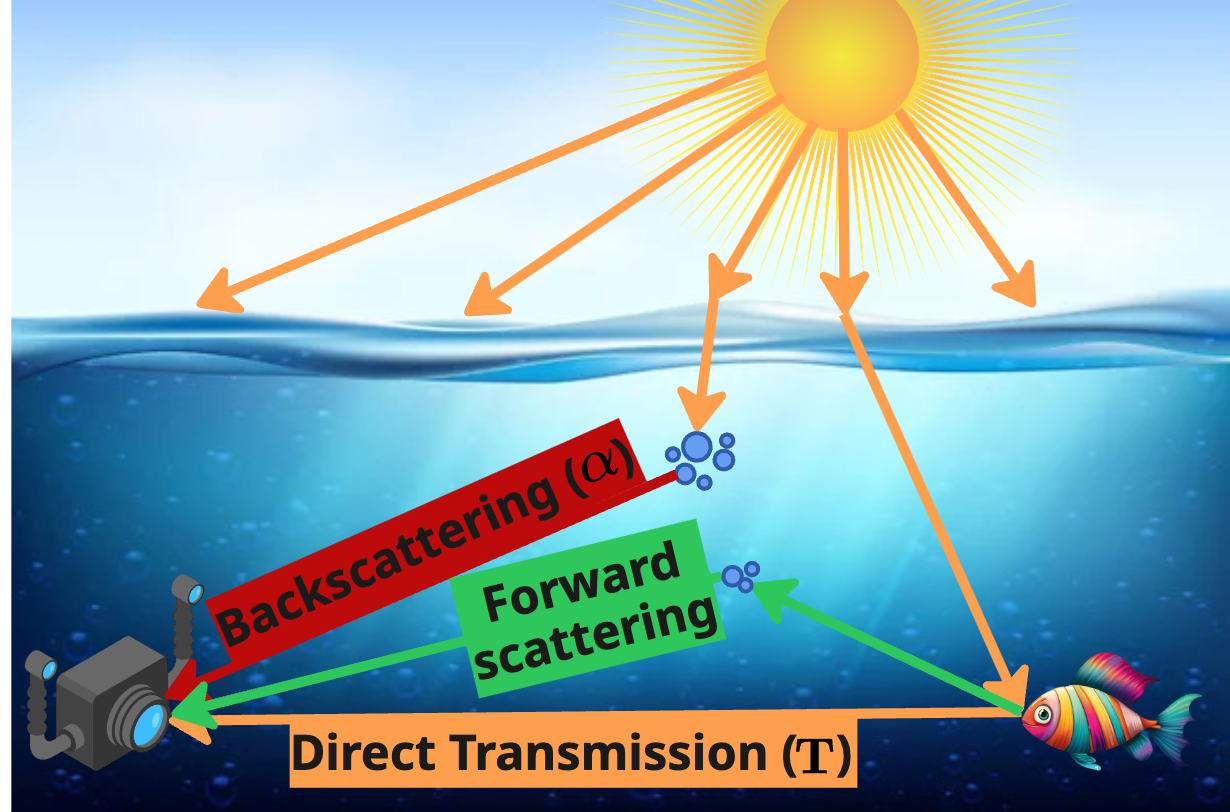}
  \caption{Jaffe-McGlamery formation model.}
  \label{fig:jmg_formation_model}
\end{figure}

Underwater images suffer from light distortions due to water absorbing light waves. 
Depth, illumination and water turbidity affect image formation, resulting in pictures with low edge definition and severe color distortion; these tend to appear blueish or greenish (\eg, samples in ~\figurename~\ref{fig:UWIN}).  Based on the Beer-Bouguer-Lambert law, the Jaffe-McGlamery model~\cite{jaffe1990computer,mcglamery1975computer} describes the degradation of underwater images by simulating light propagation through water. The Jaffe-McGlamery model decomposes the irradiance received by the camera, the light used by the camera for image formation, into direct transmission, backscattering, and forward scattering. As shown in~\figurename~\ref{fig:jmg_formation_model}, the direct transmission component is the light from the objects to the camera without scattering, attenuated exponentially with distance based on medium-specific absorption coefficients, which quantify how much the energy of the light is absorbed by the water.
The backscattering component is the ambient light reflected by water particles towards the imaging device, creating a veiling glare effect that reduces contrast. Forward scattering component is the light deflected by the suspended particles but still caught by the camera~\cite{ma2025comparative}.

At typical underwater imaging distances, forward scattering effects are negligible compared to the backscattering phenomena, so this component is omitted in favor of a simplified image degradation model defined as
\begin{equation}
\label{eq:jaffe_mcglamery_model}
   \inputimg = \realcleanoutputimg \odot \transmissionmap + \backgroundlight \odot (1 - \transmissionmap) 
\end{equation}
where $\inputimg \in \mathbb{R}^{3\times H \times W}$ denotes the degraded image, $\realcleanoutputimg \in 
\mathbb{R}^{3\times H \times W}$ is the clear image, and 
0$\backgroundlight$ is the background light, representing the backscattered light that tends to dominate in turbid conditions.
$\odot$ is the Hadamard product and $\transmissionmap \in [0,1]^{H \times W}$ is the transmission map matrix, with each element denoting the percentage of the scene radiance reaching the camera without scattering, and is defined as: 
\begin{equation}
 \transmissionmap = e^{-\nu d_{\realcleanoutputimg}} 
\end{equation}
where $\nu$ is the attenuation coefficient and $d_{\realcleanoutputimg} \in \mathbb{R}^{H\times W}$ is the distance of the object to the camera. 

By leveraging physical principles of light interaction with water, this formulation aims at modeling the fact that objects closer to the camera are less affected by scattering (\ie, $\transmissionmap$ is closer to 1) while distant objects suffer from more severe degradation (\ie, $\transmissionmap$ values move toward 0),~\ie, backscattering issues.
The physical relevance of this model motivates its use as a basis for designing a principled approach that estimates the underlying transmission and background light to reverse their effects.

\begin{figure*}[th]
  \centering
  \includegraphics[width=\linewidth]{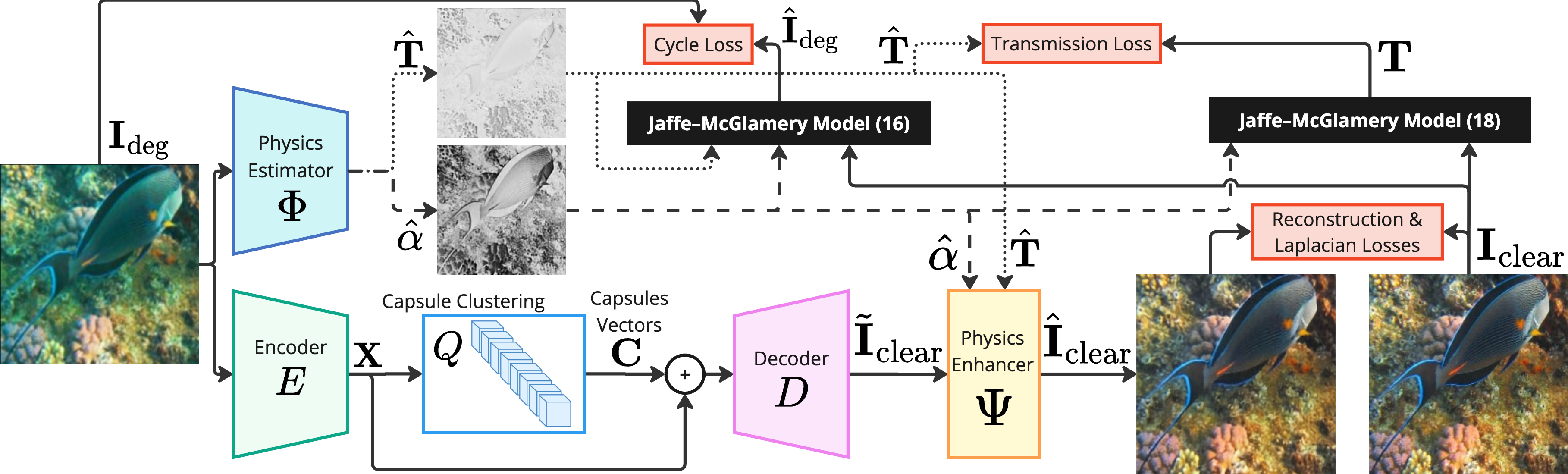}
  \caption{Proposed \myalgoname architecture with the new capsule vector latent space clusterization mechanism.}
  \label{fig:system_pipeline}
\end{figure*}

\section{Proposed Method}
\label{sec:method}
Figure \ref{fig:system_pipeline} illustrates our architecture, which consists of two parallel streams, one focused on the distribution of light (irradiance) in the image and the second dedicated to the feature extraction from the image. The encoder ($\encoder$) and physics estimator ($\physicsestimator$) start from the degraded underwater image $\inputimg$. While the $\encoder$ generates a latent image representation $\mathbf{X}$, the $\physicsestimator$ estimates the transmission map $\transmissionmapestimate$ and background light $\backgroundlightestimate$.
The latent image $\mathbf{X}$ is exploited by a capsule clustering module ($\capsuleclustering$), capturing entity-level features that are then used to augment such a representation before decoding.
The decoder ($\decoder$) works on the latent representation to generate an image-like output.
This is finally processed by the physics enhancer ($\physicsenhancer$) --exploiting the transmission map and background light estimates to reverse their effects-- emitting the enhanced image,~\ie, $\reconstructedimg \in \mathbb{R}^{3\times H \times W}$.

\subsection{Encoder ($E$)}
\label{sec:method:encoding}
Our encoder architecture is designed to extract a compact yet informative latent representation while preserving crucial spatial information. The design follows a hierarchical structure that balances computational efficiency with feature richness.

We begin by computing $\mathbf{H}_0 = \texttt{Conv2D}_{3\times3}(\inputimg)$, followed by $l\in [1,N]$ residual encoding blocks, each computing
\begin{equation}
\mathbf{H}_{l}^{\text{res}} = \resblock(\mathbf{H}_{l-1})\in \mathbb{R}^{C_{l} \times H_{l} \times W_{l}}
\end{equation}
ensuring effective information propagation through deeper layers for preserving and enhancing subtle underwater textures and colors, while mitigating vanishing gradients.
Each residual block is followed by a \convtwod halving feature resolution spatial dimensions, optimizing computational efficiency while allowing the model to capture high-level abstract features.

At the output of $N$ residual blocks, we add a self-attention mechanism followed by normalization and nonlinearity operators to further refine the extracted features and obtain the encoder output as
\begin{multline}
\mathbf{X} = \texttt{Conv2D}_{3\times 3}(\texttt{SiLU}(\texttt{GroupNorm}(\mathbf{H}_N^{\text{res}} + \\ 
\texttt{SelfAttention}(\mathbf{H}_N^{\text{res}}))) \in \mathbb{R}^{C_X \times H_X \times W_X}
\end{multline}

\subsection{Capsule Clustering ($Q$)}
\label{sec:method:cvq}
\begin{figure*}
  \centering
  \includegraphics[width=\linewidth]{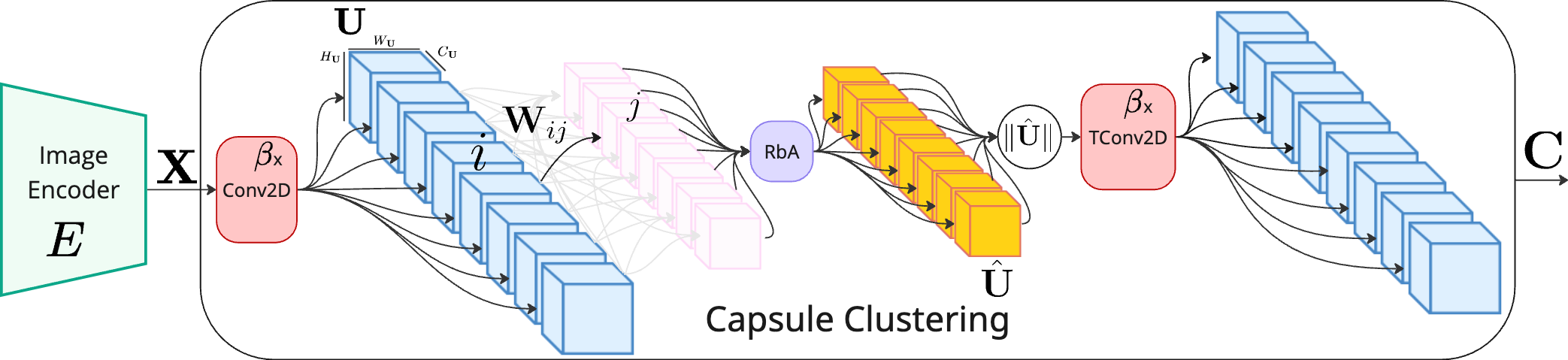}
  \caption{Proposed capsule vector clustering approach.
  It consists of a capsule layer and a convolutional transpose layer.
  The capsules extract $\mathbf{U}$ features which are clusterized by the \myRbA{ }procedure, to obtain $\mathbf{\hat{U}}$. We aggregate the matrices and upsample them by a transposed convolution layer.}
  \label{fig:capsule_clustering}
\end{figure*}

Following the encoder, we introduce a capsule network layer to model entity-level relationships within the latent image representation.
We start by processing the encoder output $\mathbf{X}$ through $\beta$ parallel convolutional layers, yielding to $\beta$ capsules, represented as $\mathbf{U} \in \mathbb{R}^{\beta \times C_{\mathbf{U}} \times H_{\mathbf{U}} \times W_{\mathbf{U}}}$ -- where $C_{\mathbf{U}}$ represents the capsule dimension,  $H_{\mathbf{U}}, W_{\mathbf{U}}$ denote the capsule grid dimension.
For each grid location, we have $\mathbf{u}_i \in \mathbb{R}^{C_{\mathbf{U}}}$ representing the output of capsule $i \in \{0, \cdots, \beta\}$.
The length of each such vector indicates the probability that a particular feature exists at a specific location, while its orientation represents the instantiation parameters (\eg,~pose, deformation, etc.)~\cite{sabour2017dynamic}.

These \textit{primary} capsules are the first level of abstraction beyond spatially preserving image features and aim at encapsulating the instantiation parameters of the detected features.
The dynamic routing algorithm then routes these capsules to higher-level capsules based on agreement, which helps in recognizing more complex structures.

Let $j\in\{0, \cdots, \gamma\}$ denote a parent capsule index.
This receives input from all $\beta$ capsules via

\begin{equation}
    \hat{\mathbf{u}}_{j|i} = \mathbf{W}_{ij} \mathbf{u}_i
\end{equation}

where $\mathbf{W}_{ij} \in \mathbb{R}^{C_{\mathbf{U}} \times C_{\hat{\mathbf{U}}}}$ defines the affine transformation matrix.
The resulting prediction vector $\hat{\mathbf{u}}_{j|i}$ estimates capsule $i$'s contribution to capsule $j$

The Routing-by-Agreement clustering algorithm starts by adaptively weighing these contributions.
We first apply a coupling coefficient $c_{ij}$ computed via softmax:
\begin{equation}
    c_{ij} = \frac{\exp(b_{ij})}{\sum_{k}\exp(b_{ik})}
\end{equation}

where $b_{ij}$ is iteratively updated via scalar product as $b_{ij} = b_{ij} + \hat{\mathbf{u}}_{j|i} \cdot \mathbf{v}_j$.

The weighted sum of prediction vectors generates:

\begin{equation}
    \mathbf{s}_j = \sum_{i} c_{ij} \hat{\mathbf{u}}_{j|i}
\end{equation}

We then apply the squashing function to obtain the activity vector $\mathbf{v}_j$:

\begin{equation}
    \mathbf{v}_j = \texttt{squash}(\mathbf{s}_j) = \frac{\Vert\mathbf{s}_j\Vert^2}{1+\Vert\mathbf{s}_j\Vert^2}\frac{\mathbf{s}_j}{\Vert\mathbf{s}_j\Vert}
\end{equation}

While $\mathbf{v}_j$ effectively captures entity presence probability, it abstracts away the precise spatial information required for accurate image reconstruction.
To preserve both entity-level and spatial information, we weight each prediction vector $\hat{\mathbf{u}}_{j|i}$ by its corresponding coupling coefficient $c_{ij}$ (from the final routing iteration) to obtain $\hat{\mathbf{U}} \in \mathbb{R}^{\beta \times C_{\hat{\mathbf{U}}}\times H_{\mathbf{U}} \times W_{\mathbf{U}}}$.
Entity presence at specific locations is then captured through $\ell_2$-norm computation over $C_{\hat{\mathbf{U}}}$, followed by a \convT~layer mapping the $\beta$ capsules to $C_{\mathbf{X}}$ feature maps, yielding the capsule vectors $\mathbf{C} \in \mathbb{R}^{C_{\mathbf{X}} \times H_{\mathbf{X}} \times W_{\mathbf{X}}}$.

Our capsule network design is motivated by the known limitations of CNNs in modeling part-to-whole relationships in image data.
CNNs excel at modeling neighboring spatial pixel relationships but lack the ability to model entity-level information --if not through a long list of layers with increased receptive fields.
By incorporating capsule networks with our novel spatial preservation mechanism via $\hat{\mathbf{U}}$, we aim at capturing both entity-level semantic information and precise spatial relationships.
This dual representation is particularly relevant for underwater image enhancement, where degradation effects like scattering and absorption have different impacts on objects based on their structure and spatial arrangement.

\subsection{Decoder ($D$)}
To reconstruct the enhanced image, we first augment the encoded image representation $\mathbf{X}$ with the capsule vectors $\mathbf{C}$.
To perform this operation as efficiently as possible, we adopted a residual approach to obtain $\hat{\mathbf{X}} = \mathbf{X} + \mathbf{C}$.
Through this, we exploit information about the presence of entities at specific locations (via $\mathbf{C}$) while also precisely modeling the pixel-level contextual information (via $\mathbf{X}$).
Since the enhancement must generate an output that preserves all the spatial details but removes the effects of underwater degradation, both such features are very relevant for reconstruction.

The decoder ($D$) increases the input ($\hat{\mathbf{X}}$) resolution to produce an intermediate enhanced estimate $\reconstructedintermediateimg \in \mathbb{R}^{3\times H\times W}$ through a sequence of 4 blocks, each consisting of a \resblock and an \upsampleblock~\cite{esser2021taming}. 

\subsection{Physics Estimator ($\Phi$)}
\label{sec:sub:physics_estimator}
In underwater imaging, the Jaffe-McGlamery physical framework has long served for describing image formation by modeling direct transmission and backscattering components.
Motivated by this principled formulation, we introduce a physics estimator, denoted by \(\Phi\), that maps the input underwater image to a two-channel tensor
\begin{equation}
    \left[\transmissionmapestimate; \backgroundlightestimate\right] = \Phi(\inputimg)
\end{equation}
where the first channel, $\transmissionmapestimate \in [0,1]^{H\times W}$, represents the estimated transmission map while $\backgroundlightestimate \in [0,1]^{H\times W}$ provides a spatially varying estimate of the background (or backscattered) light.


\subsection{Physics Enhancer ($\Psi$)}
\label{sec:sub:physics_enhancer}
The estimates of the transmission map and the backscattered light are exploited to reverse their effects on the clear image.
This is achieved by rearranging the Jaffe–McGlamery formation model while considering $\reconstructedintermediateimg$, yielding to
\begin{equation}
\reconstructedintermediateimg = \realcleanoutputimg \odot \transmissionmapestimate + \backgroundlightestimate \odot \bigl(1-\transmissionmapestimate\bigr),
\end{equation}
which, through rearrangement, can be used to obtain the final enhanced output
\begin{equation}
    \reconstructedimg \approx \realcleanoutputimg = \bigl( \reconstructedintermediateimg - \backgroundlightestimate \odot (1- \transmissionmapestimate) \bigr) \oslash \transmissionmapestimate.
    \label{eq:image_refinement}
\end{equation}
where $\oslash$ is the Hadamard division.

This parameter-free refinement step effectively removes the additive contribution of the backscattered light and normalizes the result by the transmission, thereby compensating for the degradation induced by absorption and scattering.
By directly incorporating the estimated $\transmissionmapestimate$ and $\backgroundlightestimate$ into~\eqref{eq:image_refinement}, our method ensures that the enhanced output image conforms to the physical constraints of the underwater environment.

\subsection{Optimization Objective}
Our restoration model is designed to predict an enhanced image $\reconstructedimg$, a per-pixel transmission map $\transmissionmapestimate$, and a background light map $\backgroundlightestimate$.
To ensure that our network outputs conform to this physical Jaffe–McGlamery model while yielding perceptually enhanced images, we designed four loss terms into our overall training objective.

\subsubsection{Reconstruction Loss}
To ensure spatial coherence between the noise-free ground truth (\ie, $\realcleanoutputimg$) and reconstructed image (\ie, $\reconstructedimg$), we compute:
\begin{equation}
\label{eq:reconstruction_pixel_loss}
\loss_{\text{rec}}= \| \realcleanoutputimg - \reconstructedimg \|_1
\end{equation}

\subsubsection{Laplacian Pyramid Loss}
To capture both fine details and global structures in the enhanced output, we introduce a multi resolution loss function based on the Laplacian pyramid decomposition.
This computes
\begin{align}
\label{eq:laplacian_pyramid_loss}
\loss_{\text{lap}} = \sum_{k=0}^{L-1} \omega_k \| \lambda_k(\reconstructedimg) - \lambda_k(\realcleanoutputimg) \|_1
\end{align}
where $L$ denote the pyramid levels, $\lambda_k(\cdot)$ computes the $k$-th level of the Laplacian pyramid, and $\omega_k$ represents the asscociated weight.
The $k$-th level of the Laplacian pyramid is constructed as follows:
\begin{gather}
    \label{eq:laplacian_pyramid_construction}
\lambda_k(I) = 
\begin{cases}
G_k(\realcleanoutputimg) - \upsample(G_{k+1}(\realcleanoutputimg)), & \text{if } k < L-1 \\
G_{L-1}(\realcleanoutputimg), & \text{if } k = L-1
\end{cases}
\end{gather}
where $G_k(\realcleanoutputimg)$ is the $k$-th level of the Gaussian pyramid obtained through recursive average pooling operations
\begin{equation}
    G_k(\realcleanoutputimg) =
\begin{cases}
\realcleanoutputimg , & \text{if } k = 0 \\
\avgpool(G_{k-1}(\realcleanoutputimg)) & \text{else}
\end{cases}
\end{equation}

This loss formulation enforces consistency across multiple spatial frequencies, ensuring that both local details (high frequencies affected by scattering) and global structures (low frequencies affected by color attenuation) are properly recovered. 

\subsubsection{Cycle Loss}
The cycle loss enforces consistency between the observed (\ie, degraded) image $\inputimg$ and a re-composition of the image using the physics-related network outputs.
Following (\ref{eq:jaffe_mcglamery_model}), we synthesize the degraded image $\reconstructedinputimg \in \mathbb{R}^{3\times H \times W}$ as
\begin{equation}
  \label{eq:cycle_loss:recomposition}
  \reconstructedinputimg = \realcleanoutputimg \odot \transmissionmapestimate + \backgroundlightestimate \odot \bigl(1-\transmissionmapestimate\bigr),
\end{equation}
and then define the reconstruction loss as
\begin{equation}
  \loss_{\text{cycle}} = \bigl\| \inputimg - \reconstructedinputimg \bigr\|_1.
\label{eq:cycle_loss}
\end{equation}
This term pushes the network to produce estimates of $\transmissionmapestimate$ and $\backgroundlightestimate$ that adhere to the underwater image formation model.

\subsubsection{Transmission Map Loss}
We further constrain the transmission map by deriving an expected transmission value from the formation model.
Since we have two unknowns in~\eqref{eq:jaffe_mcglamery_model}, we can rearrange~\eqref{eq:jaffe_mcglamery_model} while considering our estimate for the backscatter light $\backgroundlightestimate$
to obtain the expected transmission map
\begin{equation}
  \transmissionmap = (\inputimg - \backgroundlightestimate) \oslash (\realcleanoutputimg - \backgroundlightestimate + \varepsilon),
  \label{eq:expected_transmission}
\end{equation}
where $\varepsilon$ is a small constant to avoid division by zero.
The transmission supervision loss is then defined by
\begin{equation}
  \loss_{\text{transmission}} = \bigl\| \transmissionmap - \transmissionmapestimate \bigr\|_1.
  \label{eq:transmission_loss}
\end{equation}


\subsubsection{Optimization Loss}
The total loss function is
\begin{equation}
  \loss =  \loss_{\text{rec}} + \loss_{\text{lap}} +  \eta (\loss_{\text{cycle}} + \loss_{\text{transmission}})
  \label{eq:total_loss}
\end{equation}
where $\eta$ is the physics-related loss scaling factor.

\begin{table*}[t]
        \centering
        \caption{Quantitative comparison of \myalgoname and state-of-the-art methods on full-reference datasets ($\uparrow$ higher is better, $\downarrow$ lower is better). For each metric/dataset, the best method is in red, the second best is in blue.}
        \label{tab:full-reference_metrics}
        \footnotesize

\noindent
\makebox[\linewidth]{%
\begin{tabularx}{1.2\linewidth}{lCCCCCCCCCCCC}
\toprule 
  & \multicolumn{3}{c}{\textbf{EUVP}} & \multicolumn{3}{c}{\textbf{UFO120}} & \multicolumn{3}{c}{\textbf{LSUI}} & \multicolumn{3}{c}{\textbf{COMPLEXITY}} \\
                    & PSNR $\uparrow$   & SSIM $\uparrow$   & CLIP-IQA $\uparrow$   & PSNR $\uparrow$   & SSIM $\uparrow$   & CLIP-IQA $\uparrow$   & PSNR $\uparrow$   & SSIM $\uparrow$   & CLIP-IQA $\uparrow$   & Latency [ms] $\downarrow$  & Params [M]  $\downarrow$ & FLOPS $\downarrow$    \\
\midrule
 RGHS~\cite{huang2018rghs}               & 18.05               & 0.78                & 0.69                    & 17.70                 & 0.74                  & 0.75                      & 18.65               & 0.82                & 0.65                    & -              & -            & -        \\
 UDCP~\cite{drews2016udcp}               & 14.52               & 0.59                & 0.64                    & 14.59                 & 0.57                  & 0.72                      & 13.35               & 0.58                & 0.61                    & -              & -            & -        \\
 UIBLA~\cite{peng2017uibla}              & 18.95               & 0.74                & 0.69                    & 17.28                 & 0.66                  & 0.74                      & 18.03               & 0.74                & 0.65                    & -              & -            & -        \\
 UGAN~\cite{fabbri2018enhancing}              & 20.98               & 0.83                & 0.66                    & 20.31                 & 0.76                  & 0.73                      & 19.78               & 0.80                & 0.62                    & -              & -            & -        \\
 FUnIE-GAN~\cite{islam2020fast}         & 23.53               & 0.84                & 0.72                    & 23.76                 & 0.79                  & 0.75                      & -                   & -                   & -                       & -              & -            & -        \\
 Cluie-Net~\cite{li2023cluienet}          & 18.90               & 0.78                & 0.67                    & 18.65                 & 0.74                  & \tblsecond{0.78}                      & 18.88               & 0.80                & 0.66                    & 6.39           & 13.40        & 61.98G  \\
 DeepSESR~\cite{islam2020simultaneous}           & 24.22               & 0.85                & 0.60                    & 24.02                 & 0.81                  & 0.76                      & -                   & -                   & -                       & -              & -            & -        \\
 TWIN~\cite{liu2022twin}               & 18.91               & 0.79                & 0.64                    & 18.48                 & 0.74                  & 0.75                      & 20.11               & 0.81                & 0.64                    & 13.60          & 11.38        & 198.57G \\
 UShape-Transformer~\cite{peng2023ushape-lsuidataset} & 27.59               & 0.88                & 0.64                    & 23.51                 & 0.80                  & 0.73                      & 23.64               & 0.84                & 0.64                    & 49.08          & 31.59        & 52.24G  \\
 Spectroformer~\cite{deng2018hyperspectral}      & 18.70               & 0.79                & 0.69                    & 18.29                 & 0.74                  & 0.77                      & 20.41               & 0.81                & 0.69                    & 47.81          & 2.43         & 35.63G  \\
 CE-VAE~\cite{pucci2025cevae}             & 27.75               & 0.88                & 0.69                    & 25.26                 & 0.82                  & 0.77                      & 25.32               & 0.86                & 0.66                    & 62.33          & 83.44        & 473.77G \\
 DM-Underwater~\cite{tang2023dmwater}      & 26.73               & 0.88                & 0.66                    & 25.36                 & 0.83                  & 0.76                      & \tblsecond{27.77}    & 0.90                & 0.64                    & 229.82         & 18.34        & 1.34T   \\
 WF-Diff~\cite{zhao2024wfdiff}            & 26.94               & \tblsecond{0.89}                & 0.50                    & 25.64                 & \tblsecond{0.84}                  & 0.62                      & 24.95               & 0.88                & 0.50                    & 393.21         & 100.55       & 2.41T   \\
 PA-Diff~\cite{zhao2024padiff}            & \tblsecond{28.47}    & \tblfirst{0.91}      & \tblsecond{0.76}          & \tblsecond{26.48}      & \tblfirst{0.86}        & \tblfirst{0.85}            & 26.28               & \tblfirst{0.91}     & \tblsecond{0.71}         & 351.73         & 56.12        & 3.65T   \\
\rowcolor{lightgray!30} \myalgoname          & \tblfirst{28.91}     & \tblfirst{0.91}     & \tblfirst{0.77}         & \tblfirst{26.53}       & \tblfirst{0.86}       & \tblfirst{0.85}           & \tblfirst{27.81}     & \tblfirst{0.91}      & \tblfirst{0.72}          & 80.71          & 92.42        & 900.57G \\
\bottomrule
\end{tabularx}
}
\end{table*}

\section{Experimental Results}
\label{sec:exp}

\subsection{Datasets}
\label{sec:datasets}
We validate our method on six benchmark datasets to assess its generalization across diverse underwater conditions.
To perform a comparison between the enhanced image and the available ground truth, we considered the following full-reference datasets:
(i) the LSUI-L400 dataset~\cite{peng2023ushape-lsuidataset} comes with images featuring different water types, lighting conditions, and target categories\footnote{
The evaluation considers the Test-L 400 split proposed in~\cite{peng2023ushape-lsuidataset}.};
(ii) the EUVP dataset\cite{islam2020fast} comprises 1970 validation image samples of varying quality;
and 
(iii) the UFO-120 dataset~\cite{islam2020simultaneous} contains 120 full-reference images collected from oceanic explorations across multiple locations and water types.

To validate our approach in a broader context, we extended our model analysis to three non-reference datasets:
(i) the UCCS dataset~\cite{8949763} consists of 300 images of marine organisms/environments specifically acquired to evaluate color cast correction in underwater image enhancement;
(ii) the U45~\cite{li2019fusion} and (iii) SQUID~\cite{berman2020underwater} datasets contain 45 and 57 raw underwater images  respectively. Images show severe color casts, low contrast, and haze degradations.

\subsection{Metrics}
We followed recent works~\cite{li2019fusion,liu2022twin,peng2023ushape-lsuidataset,khan2024spectroformer}, and assessed our model performance considering the Peak Signal-to-Noise Ratio (PSNR), the Structural Similarity (SSIM)~\cite{wang2004ssim}, and the Learned Perceptual Image Patch Similarity (LPIPS)~\cite{zhang2018unreasonable} for full-reference datasets. 

For non-reference datasets, we considered the Underwater Color Image Quality Evaluation Metric (UCIQE)~\cite{7300447}, the Underwater Image Quality Measure (UIQM)~\cite{7305804}, and the CLIP-IQA Score ~\cite{wang2023clipiqa}.

\subsection{Implementation Details}
For a fair comparison with existing methods~\cite{tang2023dmwater,zhao2024padiff,zhao2024wfdiff}, we run the experimental evaluation with random cropped and horizontal flipped $\inputimg \in \mathbb{R}^{3\times H=256\times W=256}$.
Our encoder ($E$) has $N=4$ residual blocks that emit $\mathbf{X}\in\mathbb{R}^{256\times 16\times 16}$.
The capsule clustering ($Q$) has $\beta=32$ capsules yielding to $\mathbf{U} \in \mathbb{R}^{32\times 16 \times9 \times9}$.
The \myRbA{ } algorithm is run for 3 iterations, with $\beta = 32$ to obtain $\hat{\mathbf{U}}\in \mathbb{R}^{32\times 16\times 16\times 16}$. 
The normalization and following transposed convolution layers output $\mathbf{C} \in \mathbb{R}^{256\times 16\times 16}$.
We train our model for $500$ epochs, with a batch size of $32$ using the AdamW optimizer with a learning rate of $4.5e^{-6}$ on the LSUI Train-L dataset~\cite{peng2023ushape-lsuidataset}.
We set $L=3$ pyramid levels controlling the weights $\omega_k = 1/2^k$ (with $k \in \{1, \cdots, L-1\}$ and used $\eta=0.0001$.

\subsection{State-of-the-art Comparison}
\label{sec:exp:sota}
We compare the performance of our \myalgoname model with existing traditional methods like RGHS~\cite{huang2018rghs}, UDCP~\cite{drews2016udcp}, and UIBLA~\cite{peng2017uibla} as well as state-of-the-art machine learning-based works including UShape-Transformer~\cite{peng2023ushape-lsuidataset}, 
Spectroformer~\cite{khan2024spectroformer}, DM-Water~\cite{tang2023dmwater}, CEVAE~\cite{pucci2025cevae}, WF-Diff~\cite{zhao2024wfdiff}, PA-Diff~\cite{zhao2024padiff}.
We report on the results published in the corresponding papers or by running the publicly available codes using the same training data.

\begin{table*}[t]
        \centering
        \caption{Quantitative comparison of \myalgoname and state-of-the-art methods on non-reference datasets ($\uparrow$ higher is better, $\downarrow$ lower is better). For each metric/dataset, the best method is in red, the second best is in blue.}
        \label{tab:non-reference_metrics}
        \footnotesize

        \noindent
\makebox[\linewidth]{%
\begin{tabularx}{1.2\linewidth}{l HCCHC HCCHC HCCHC CCC}
\toprule 
  & \multicolumn{5}{c}{\textbf{U45}} & \multicolumn{5}{c}{\textbf{SQUID}} & \multicolumn{5}{c}{\textbf{UCCS}} & \multicolumn{3}{c}{\textbf{COMPLEXITY}} \\
                    & NIQE $\downarrow$   & UIQM $\uparrow$   & UCIQE $\uparrow$   & BRISQUE $\downarrow$   & CLIP-IQA $\uparrow$   & NIQE $\downarrow$   & UIQM $\uparrow$   & UCIQE $\uparrow$   & BRISQUE $\downarrow$   & CLIP-IQA $\uparrow$   & NIQE $\downarrow$   & UIQM $\uparrow$   & UCIQE $\uparrow$   & BRISQUE $\downarrow$   & CLIP-IQA $\uparrow$   & Latency [ms] $\downarrow$  & Params [M] $\downarrow$   & FLOPS$\downarrow$    \\
\midrule
 UDCP~\cite{drews2016udcp}               & 4.83                 & 2.09               & 0.59                & 8.49                    & 0.79                   & 9.39                   & 1.27                 & \tblsecond{0.56}        & 3.30                      & 0.76           & 5.55                  & 2.17                & 0.55                 & 19.48                    & 0.38                    & -              & -            & -        \\
 UGAN~\cite{fabbri2018enhancing}               & 6.56                 & 3.04               & 0.55                & 22.93                   & 0.77                   & 8.81                   & 2.38       & 0.52                  & 14.18                     & 0.46                     & 6.85                  & 2.84                & 0.51                 & 25.57                    & 0.36                    & -              & -            & -        \\
 Cluie-Net~\cite{li2023cluienet}          & 4.41                 & 3.19               & 0.59                & 8.62                    & 0.80                   & \tblsecond{7.13}        & 2.12                 & 0.51                  & 15.15                     & 0.80                     & 5.19                  & 3.02                & 0.55                 & 22.20                    & 0.42         & 7.43           & 13.40        & 61.98G  \\
 UShape-Transformer~\cite{peng2023ushape-lsuidataset} & 4.92                 & 3.11               & 0.59                & \tblsecond{13.38}        & 0.67                   & 8.33                   & 2.21                 & 0.54                  & 31.33                     & 0.68                     & \tblsecond{4.69}       & 3.13                & 0.56                 & \tblfirst{18.16}          & 0.47                    & 49.56          & 31.59        & 52.24G  \\
 Spectroformer~\cite{khan2024spectroformer}      & \tblsecond{4.22}      & 3.21               & \tblfirst{0.61}      & 8.20                    & \tblfirst{0.85}         & \tblfirst{6.56}         & \tblfirst{2.45}      & \tblsecond{0.56}                  & \tblfirst{12.61}           & 0.81          & 4.80                  & \tblsecond{3.20}      & 0.55                 & 22.07                    & 0.46          & 47.62          & 2.43         & 35.63G  \\
 CE-VAE~\cite{pucci2025cevae}             & 5.79                 & 3.18               & 0.59                & 19.55                   & 0.77                   & 8.97                   & 2.35                 & \tblsecond{0.56}       & 19.63                     & 0.63                     & 5.31                  & \tblsecond{3.20}                & 0.56                 & 24.92                    & 0.46                    & 60.00          & 83.44        & 473.77G \\
 DM-Underwater~\cite{tang2023dmwater}      & 4.58                 & \tblfirst{3.23}     & 0.59                & 6.48                    & 0.80                   & 7.44                   & 2.31                 & 0.55                  & \tblsecond{13.92}          & \tblsecond{0.82}                     & 4.92                  & 3.19     & 0.56      & \tblsecond{18.18}         & \tblfirst{0.49}                    & 229.11         & 18.34        & 1.34T   \\
 WF-Diff~\cite{zhao2024wfdiff}            & \tblfirst{39.61}      & 3.05               & 0.56                & 43.62                   & 0.55                   & -                    & 2.18                 & 0.50                  & 84.51                     & 0.45                     & \tblfirst{23.04}       & 3.06                & 0.55                 & 30.46                    & 0.24                    & 375.61         & 100.55       & 2.41T   \\
 PA-Diff~\cite{zhao2024padiff}            & 4.57                 & 3.09               & 0.58                & 8.92                    & 0.82        & 7.17                   & 2.05                 & 0.54                  & 22.44                     & 0.82                     & 5.03                  & 3.12                & 0.56                 & 20.72                    & \tblsecond{0.48}                    & 350.34         & 56.12        & 3.65T   \\
 \rowcolor{lightgray!30}
 \myalgoname         & 4.61                 & \tblsecond{3.22}    & \tblfirst{0.61}     & \tblfirst{10.05}         & \tblsecond{0.83}                   & 7.69                   & \tblsecond{2.42}                 & \tblfirst{0.57}                  & 3.12                      & \tblfirst{0.83}                     & 5.27                  & \tblfirst{3.21}                & \tblfirst{0.57}       & 25.15                    & \tblsecond{0.48}                    & 80.82          & 92.42        & 900.57G \\
\bottomrule
\end{tabularx}
}
\end{table*}

\subsubsection{Full-reference datasets.}
Table~\ref{tab:full-reference_metrics} shows that across diverse underwater datasets, our method consistently showcases state-of-the-art underwater image enhancement performance while requiring substantially lower computational resources.
On the EUVP dataset,~\myalgoname has the highest PSNR with a score of  28.91 dB, while obtaining comparable performance to the previous best existing model (namely PA-Diff~\cite{zhao2024padiff}). Similarly, on UFO120 and LSUI datasets, our approach achieves the highest PSNRs and obtains similar SSIM and CLIP-IQA performance with the best competing methods.
Compared to the top-performing existing methods, our approach has a computational cost of 900.57 GFLOPS, yielding to 80.71 ms of latency.
This represents a $4.3\times$ reduction in computational cost compared to PA-Diff (351.73 ms, 3.65 TFLOPS).
These results demonstrate that \myalgoname precisely reconstructs the spatial relation between entities with great details under different water types, locations, lighting conditions, and multiple targets --effectively balancing enhancement quality and computational efficiency.

\subsubsection{Non-reference datasets.}
Table \ref{tab:non-reference_metrics} presents a quantitative comparison between our \myalgoname method and state-of-the-art approaches on non-reference underwater image datasets.
Results show that we score at the top of the leaderboard for 5 out of 9 metrics and have the second-best result for the remaining 4. 
Specifically, on the U45 dataset, we have the best UCIQE (0.61), second-best in UIQM (3.22) and CLIP-IQA (0.83). For SQUID, our approach achieves top performance in UCIQE (0.57) and CLIP-IQA (0.83), while securing second place in UIQM (2.42).
On UCCS, \myalgoname obtains the highest UIQM (3.21) and UCIQE (0.57) scores, with competitive CLIP-IQA (0.48).

\subsection{Ablation Study}
\label{sec:exp:ablation}
Through the ablation study, we want to answer different questions that would help us understand the importance of each proposed component of our architecture.

\begin{table*}[t]
        \centering
\caption{Ablation study comparing different capsule integration mechanisms for our \myalgoname approach on the three considered full-reference datasets.
We evaluate three strategies: direct capsule usage ($\hat{\mathbf{X}} = \mathbf{C}$), feature concatenation ($\hat{\mathbf{X}} = \conv_{1\times 1}(\concat(\mathbf{X}, \mathbf{C}))$), and residual connection ($\hat{\mathbf{X}} = \mathbf{X} + \mathbf{C}$).}
        \label{tab:capsule_fusion_alternatives}
        \footnotesize
\begin{tabularx}{\linewidth}{lCCCCCCCCCHHH}
\toprule 
  & \multicolumn{3}{c}{\textbf{EUVP}} & \multicolumn{3}{c}{\textbf{UFO120}} & \multicolumn{3}{c}{\textbf{LSUI}} & \multicolumn{3}{H}{\textbf{COMPLEXITY}} \\
                       & PSNR $\uparrow$   & SSIM $\uparrow$   & CLIP-IQA $\uparrow$   & PSNR $\uparrow$   & SSIM $\uparrow$   & CLIP-IQA $\uparrow$   & PSNR $\uparrow$   & SSIM $\uparrow$   & CLIP-IQA $\uparrow$   &   Latency [ms] &   Params [M] & FLOPS   \\
\midrule
 $\hat{\mathbf{X}} = \mathbf{C}$ & 27.65               & 0.88               & 0.64                 & 25.44                 & 0.82                  & 0.70                      & 25.92               & 0.87                & 0.61                    &          81.42 &        92.42 & 900.57G \\
 $\hat{\mathbf{X}} = \conv_{1\times 1}(\concat(\mathbf{X}, \mathbf{C}))$         & \tblsecond{28.87}    & \tblsecond{0.90}      & \tblsecond{0.76}          & \tblsecond{26.47}      & \tblsecond{0.84}       & \tblsecond{0.83}           & \tblsecond{27.60}    & \tblsecond{0.89}      & \tblsecond{0.71}         &          85.78 &        92.42 & 900.57G \\
 \rowcolor{lightgray!30} $\hat{\mathbf{X}} = \mathbf{X} + \mathbf{C}$ (\myalgoname)           & \tblfirst{28.91}     & \tblfirst{0.91}     & \tblfirst{0.77}         & \tblfirst{26.53}       & \tblfirst{0.86}        & \tblfirst{0.85}            & \tblfirst{27.81}     & \tblfirst{0.91}     & \tblfirst{0.72}          &          80.71 &        92.42 & 900.57G \\
\bottomrule
\end{tabularx}
\end{table*}
\subsubsection{Capsule Latent Space Modeling.}
In Table~\ref{tab:capsule_fusion_alternatives} we analyze different fusion mechanisms between capsules $\mathbf{C}$
and their input features $\mathbf{X}$ in our \myalgoname approach.
The results demonstrate that the residual connection approach (\ie, $\hat{\mathbf{X}} = \mathbf{X} + \mathbf{C}$) consistently outperforms alternative fusion strategies.
Using capsules directly as latent representations ($\hat{\mathbf{X}} = \mathbf{C}$) yields the lowest performance across all datasets, with PSNR values of 27.65 dB, 25.44 dB, and 25.92 dB on EUVP, UFO120, and LSUI datasets, respectively.
The concatenation approach (\ie, $\hat{\mathbf{X}} = \conv_{1\times 1}(\concat(\mathbf{X}, \mathbf{C}))$) shows improved performance but introduces additional computational overhead.
Our adopted residual mechanism achieves the best performance across all metrics, demonstrating that the simple additive integration enables more effective feature preservation and enhancement, leading to superior reconstruction quality without increasing computational complexity.

\subsubsection{How Relevant is the Physics Enhancer?}
\begin{figure*}[t] 
    \centering
    \begin{subfigure}[b]{0.48\textwidth} 
        \centering
        \includegraphics[width=\textwidth]{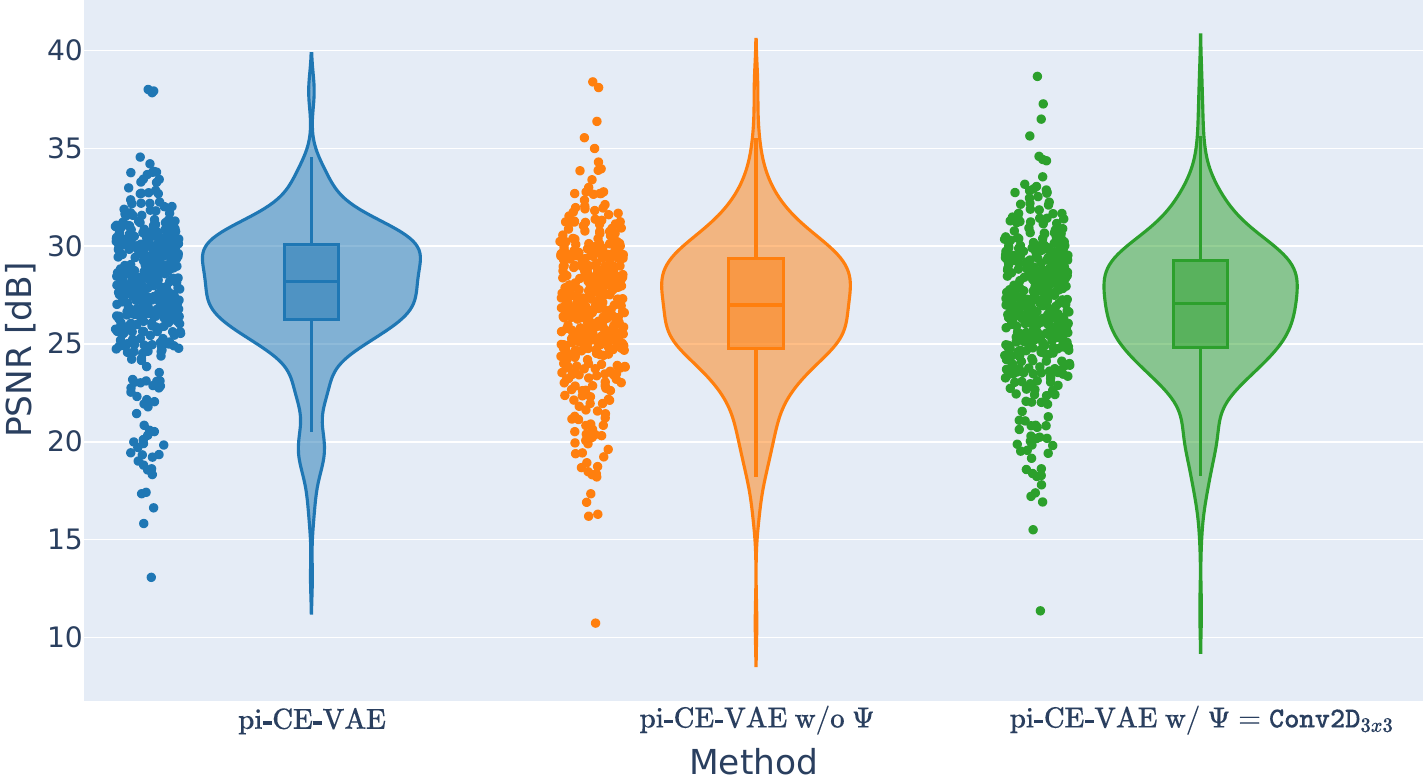}
        \caption{EUVP}
        \label{fig:subfiga}
    \end{subfigure}
    \hfill 
    \begin{subfigure}[b]{0.48\textwidth} 
        \centering
        \includegraphics[width=\textwidth]{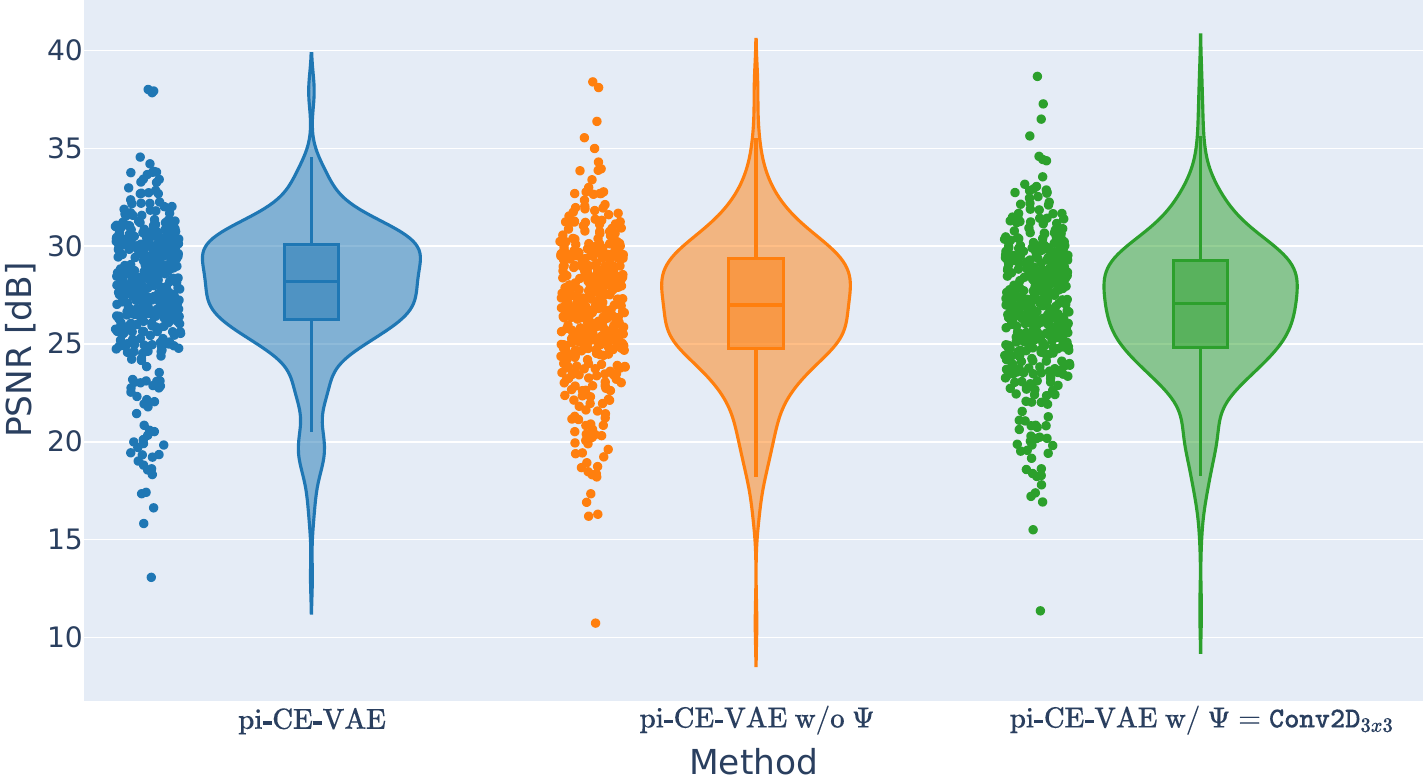} 
        \caption{UFO120}
        \label{fig:subfigb}
    \end{subfigure}
    \hfill 
    \begin{subfigure}[b]{0.48\textwidth} 
        \centering
        \includegraphics[width=\textwidth]{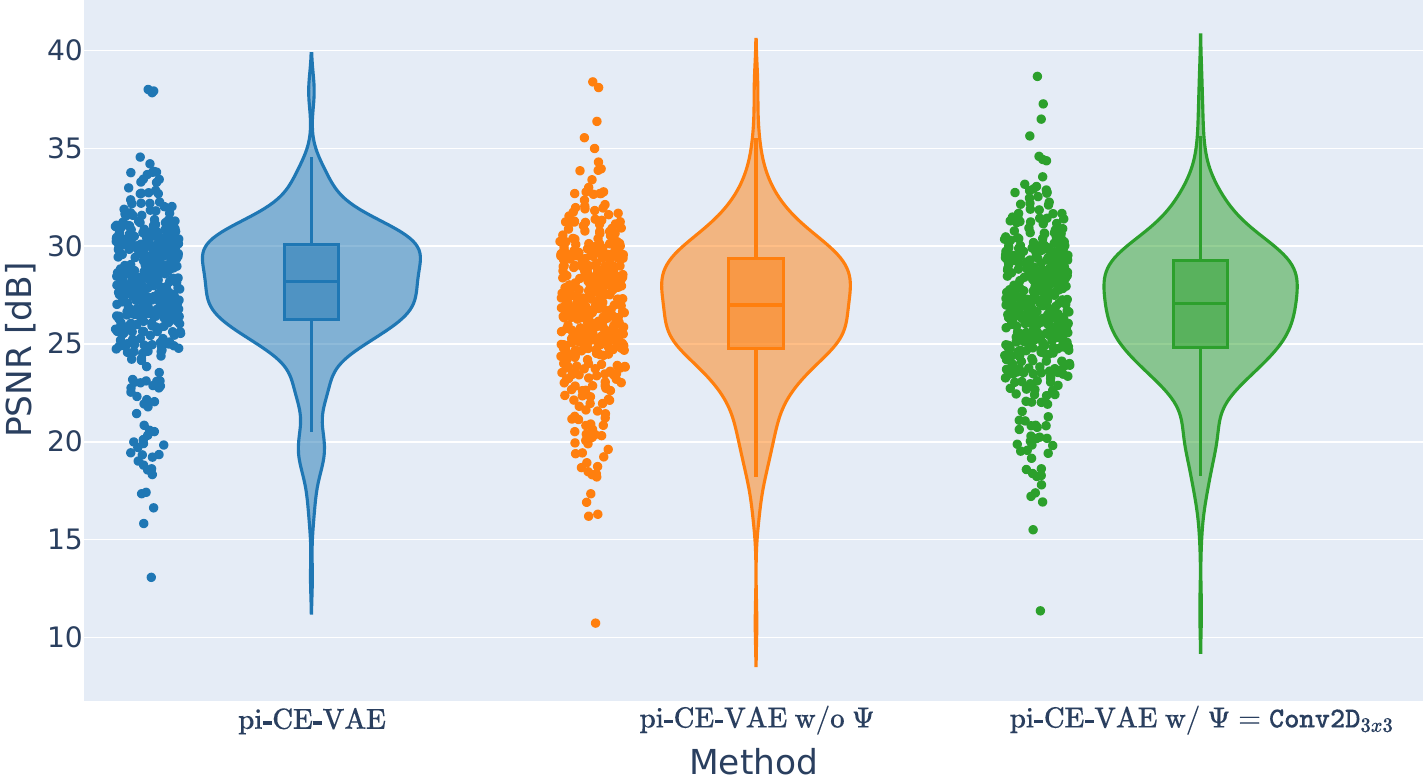} 
        \caption{LSUI}
        \label{fig:subfigc}
    \end{subfigure}
    \caption{Ablation results for different configurations of the physics enhancer ($\Psi$) computed for the (a) EUVP, (b) UFO120, and (c) LSUI datasets.}
    \label{fig:ablation_physics_enhancer}
\end{figure*}
In~\figurename~\ref{fig:ablation_physics_enhancer} we analyze the performance of our method without the physics enhancer (\ie, \myalgoname w/o $\Psi$) or with the physics enhancer replaced by a $\convtwod_{3\times 3}$ layer. 
The violin plots computed for the three full-reference datasets demonstrate that considering the physics of the Jaffe-McGlamery formation model --with estimates of the backscattering and transmission map-- consistently outperform other variants (\ie, \myalgoname has higher average PSNRs and samples are more distributed towards large PSNR values).
These results show the importance of considering such a formation model that, without extra learnable parameters, performs better than a learnable convolutional layer.

\subsubsection{How Relevant Are the Loss Terms?}
\begin{table*}[t]                                                                                                                          \centering
\caption{Ablation study on loss function components for our \myalgoname method.
The study evaluates the contribution of cycle loss ($\loss_{\text{cycle}}$), transmission map loss ($\loss_{\text{transmission}}$), Laplacian pyramid loss ($\loss_{\text{lap}}$), and reconstruction loss ($\loss_{\text{rec}}$) across different configurations.}
\label{tab:ablation_loss}

\footnotesize
\begin{tabularx}{\linewidth}{H CCCC CCC CCC CCC HHH}
\toprule

   &  $\loss_{\text{cycle}}$ &  $\loss_{\text{transmission}}$ & $\loss_{\text{lap}}$  &  $\loss_{\text{rec}}$ & 
   \multicolumn{3}{c}{\textbf{EUVP}} &
   \multicolumn{3}{c}{\textbf{UFO120}} & 
   \multicolumn{3}{c}{\textbf{LSUI}} & 
   \multicolumn{3}{H}{\textbf{COMPLEXITY}} \\
    
    & & & & & PSNR $\uparrow$   & SSIM $\uparrow$   & CLIP-IQA $\uparrow$   & PSNR $\uparrow$   & SSIM $\uparrow$   & CLIP-IQA $\uparrow$   & PSNR $\uparrow$   & SSIM $\uparrow$   & CLIP-IQA $\uparrow$   &  
    Latency [ms] &   Params [M] & FLOPS   \\

\midrule

 CE-VAE-v2-loss-no-physics-lapl   & & & \tick (3) &  & 28.26               & 0.90                & 0.70                    & 26.29                 & 0.84                  & 0.78                      & 27.37               & 0.90                & 0.67
           &          85.15 &        92.42 & 900.57G \\  

CE-VAE-v2-loss-no-physics-l1  & &  & & \tick    & 28.47               & 0.90                & 0.70                    & 26.33                 & 0.84                  & 0.77                      & 27.44               & 0.90                & 0.66                    &          64.75 &        92.42 & 900.57G \\

 CE-VAE-v2-loss-no-physics-l1+lapl & & & \tick (3) & \tick  & {28.77}     & 0.90                & \tblsecond{0.76}         & {26.36}      & 0.84       & {0.82}           & {27.56}     & \tblfirst{0.91}      & \tblsecond{0.71}
                  &          82.95 &        92.42 & 900.57G \\

 CE-VAE-v2-loss-no-cycle           & & \tick & \tick  (3)&  \tick & 28.56               & 0.90                & 0.71                    & 26.39                 & 0.84                  & 0.78                      & 27.30               & 0.90                & 0.67
           &          82.14 &        92.42 & 900.57G \\

 CE-VAE-v2-loss-no-tx     & \tick & & \tick  (3)& \tick          & \tblsecond{28.82}              & 0.90                & \tblsecond{0.76}                    & 26.45                 & 0.84                  & 0.82                      & {27.65}    & 0.90                & \tblsecond{0.71}
            &          79.28 &        92.42 & 900.57G \\

\hline

 CE-VAE-v2-loss-lapl-levels-2  & \tick & \tick & \tick (2) & \tick    & 28.67               & {0.90}     & \tblsecond{0.76}          & 26.41                 & 0.84                  & \tblsecond{0.83}                      & 27.71               & 0.90                & \tblsecond{0.71}         &         102.13 &        92.42 & 900.57G \\   

CE-VAE-v2-loss-lapl-levels-4    & \tick & \tick & \tick (4) & \tick  & 28.46               & {0.90}      & 0.73                    & 26.36                 & 0.84                  & 0.80                      & 27.50               & {0.90}     & 0.69                    &          80.61 &        92.42 & 900.57G \\

CE-VAE-v2-loss-lapl-w111   & \tick & \tick & \tick ($\overline{3}$) & \tick        & 28.79               & 0.90                & 0.74                    & \tblsecond{26.51}       & 0.84                  & 0.81                      & \tblsecond{27.72}               & 0.90                & 0.69                    &          95.08 &        92.42 & 900.57G \\

\rowcolor{lightgray!30} pi-CE-VAE        & \tick & \tick & \tick (3) & \tick                  & \tblfirst{28.91}    & \tblfirst{0.91}                & \tblfirst{0.77}                    & \tblfirst{26.53}                 & \tblfirst{0.86}       & \tblfirst{0.85}            & \tblfirst{27.81}               & \tblfirst{0.91}                & \tblfirst{
0.72}          &          80.71 &        92.42 & 900.57G \\

\bottomrule
\end{tabularx}
\end{table*}
Table~\ref{tab:ablation_loss} presents an ablation study examining the contribution of different loss components in our \myalgoname approach.
The results demonstrate the cumulative benefits of incorporating multiple loss terms for enhanced performance.

We start by analyzing a baseline model using the reconstruction loss alone.
This achieves a PSNR of $28.47$ dB, $26.33$ dB, and $27.44$ dB for the three full-reference EUVP/UFO120/LSUI datasets, respectively.
Considering only the Laplacian pyramid loss yields a slight degradation, but combining yields improvements (\ie, $+0.3$ dB, $+0.03$ dB, $+0.12$ dB).
Incorporating the transmission loss reduces such a gain, while adding the cycle consistency loss provides complementary improvements.

The Laplacian pyramid configuration analysis (last 4 rows) demonstrates the effectiveness of our design choices.
While uniform weighting across 3 levels ($\overline{3}$ means $\omega_k = 1$) achieves strong results (28.79 dB, 26.51 dB, 27.72 dB), our exponential weighting scheme ($\omega = [1/2^i]_{i=0}^{L-1}$) with 3-level pyramids proves optimal.
Such a 3-level configuration (our complete \myalgoname formulation) outperforms both 2-level and 4-level alternatives, indicating the appropriate balance between multi-scale representation and computational efficiency.


\subsection{Qualitative Analysis}
\label{sec:qualitative_analysis}
\begin{figure*}[t]
  \centering
  \includegraphics[width=\linewidth]{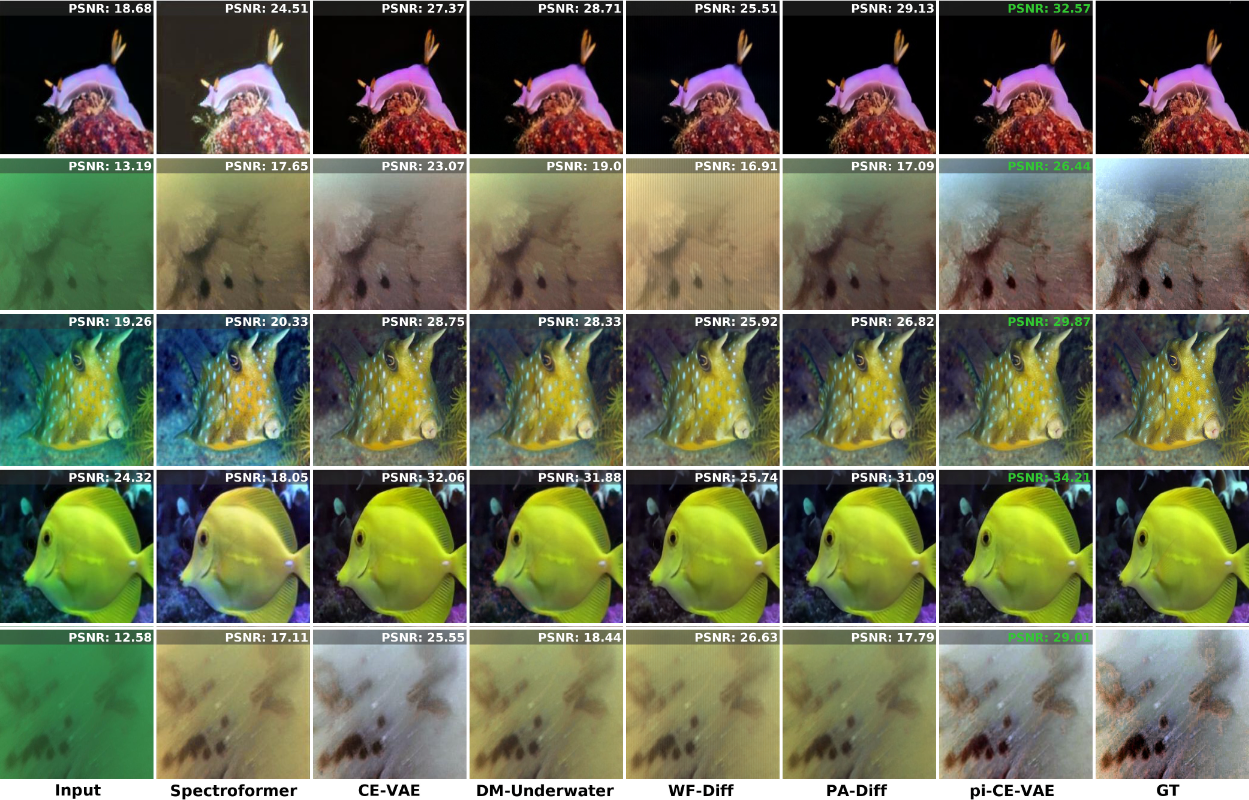}
  \caption{Enhanced images comparison on five random samples taken from the validation set of the three considered full-reference datasets.} 
  \label{fig:sota_qualitative_comparison}
\end{figure*}
To qualitatively evaluate the performance of our method, we computed the results in~\figurename~\ref{fig:sota_qualitative_comparison}.
This compares the results of our methods with the top-5 existing methods on 5 random images taken from the three full-reference datesets.
Qualitative results support the numerical performance demonstrated by our approach providing neat and realistic color restorations for different underwater challenges.

\section{Conclusions}
\label{sec:conclusion}
We have presented a novel dual-stream architecture that achieves state-of-the-art underwater image enhancement by explicitly integrating the Jaffe-McGlamery physical model with capsule clustering-based feature representation. Our physics estimator predicts transmission maps and spatially-varying background light while a parallel stream captures entity-level features, enabling parameter-free enhancement that respects physical constraints while preserving semantic structures.
We also introduced a novel optimization objective combining a multi-scale image reconstruction term with physically-related terms to ensure both adherence with the image formation model and perceptual quality. 
Evaluation across six benchmarks demonstrates that our physics-informed approach establishes new performance benchmarks with consistent and significant improvements over existing methods --while also being more computationally efficient.

\section*{Data Availability}
The LSUI dataset~\cite{peng2023ushape-lsuidataset} is available at \url{https://bianlab.github.io/data.html}.
The EUVP dataset~\cite{islam2020fast} is available at \url{https://irvlab.cs.umn.edu/resources/euvp-dataset}.
The UFO-120 dataset~\cite{islam2020simultaneous} is available at~\url{https://irvlab.cs.umn.edu/resources/ufo-120-dataset}.
The UCCS dataset~\cite{8949763} is available at~\url{https://github.com/dlut-dimt/Realworld-Underwater-Image-Enhancement-RUIE-Benchmark}.
The U45 dataset~\cite{li2019fusion} is available at~\url{https://github.com/IPNUISTlegal/underwater-test-dataset-U45-}.
The SQUID dataset~\cite{berman2020underwater} is available at~\url{https://csms.haifa.ac.il/profiles/tTreibitz/datasets/ambient_forwardlooking/index.html}.
The source code developed to train and evaluate the proposed approach \textit{will} be available at \url{https://github.com/iN1k1/}.


\section*{Funding}
Funding information - not applicable.

\bibliography{manuscript}

\end{document}